\documentclass{ieeetj}
\usepackage{cite}
\usepackage{amsmath,amssymb,amsfonts}
\usepackage{algcompatible}
\usepackage{algorithm}
\usepackage{graphicx,color}
\usepackage{textcomp}
\usepackage[table]{xcolor}
\usepackage{hyperref}
\hypersetup{hidelinks=true}
\usepackage{multirow}
\usepackage[caption=false,labelfont=sf,textfont=sf]{subfig}
\usepackage{bm}
\usepackage{dsfont}
\usepackage{url}
\def\BibTeX{{\rm B\kern-.05em{\sc i\kern-.025em b}\kern-.08em
    T\kern-.1667em\lower.7ex\hbox{E}\kern-.125emX}}
\AtBeginDocument{\definecolor{tmlcncolor}{cmyk}{0.93,0.59,0.15,0.02}\definecolor{NavyBlue}{RGB}{0,86,125}}

\def\OJlogo{\vspace{-4pt}$<$Society logo(s) and publication title will appear here.$>$}
\def\seclogo{\vspace{10pt}$<$Society logo(s) and publication title will appear here.$>$}

\def\authorrefmark#1{\ensuremath{^{\textbf{#1}}}}

\DeclareMathOperator{\argmax}{arg\,max}

\DeclareMathOperator{\len}{len}
\DeclareMathOperator{\ngram}{ngram}
\DeclareMathOperator{\autobleu}{auto-BLEU}
\DeclareMathOperator{\editdist}{edit-distance}

\algblock{PARFOR}{ENDPARFOR}
\algnewcommand\algorithmicendparfor{\textbf{end for}}
\algrenewtext{PARFOR}[1]{\textbf{for}\ #1\ \textbf{do in parallel}}
\algrenewtext{ENDPARFOR}{\textbf{end for}}

\begin{document}
\receiveddate{XX Month, XXXX}
\reviseddate{XX Month, XXXX}
\accepteddate{XX Month, XXXX}
\publisheddate{XX Month, XXXX}
\currentdate{XX Month, XXXX}
\doiinfo{XXXX.2022.1234567}

\markboth{SPEAKER-DISENTANGLED CHUNK-WISE REGRESSION FOR SYLLABIC TOKENIZATION}{KOMATSU {ET AL.}}

\title{Speaker-Disentangled Chunk-Wise Regression for Syllabic Tokenization}

\author{RYOTA KOMATSU\authorrefmark{1}, KOTA KAWAKITA\authorrefmark{1}, TAKUMA OKAMOTO\authorrefmark{2} (Member, IEEE),\\ 
AND TAKAHIRO SHINOZAKI\authorrefmark{1} (Member, IEEE)}
\affil{Institute of Science Tokyo, Meguro, Tokyo 152-8550, Japan}
\affil{National Institute of Information and Communications Technology, Kyoto 619-0289, Japan}
\corresp{Corresponding author: Ryota Komatsu (email: komatsu.r.ab@m.titech.ac.jp).}
\authornote{This work was supported in part by JTEKT Corporation and in part by JSPS KAKENHI under Grant JP22K12069.}

\begin{abstract}
Unsupervised syllabic tokenization aims to learn discrete syllabic tokens that capture latent linguistic content-related structure from raw speech. Recent syllabic tokenization methods employ teacher-student distillation of the pretrained HuBERT to organize latent speech frame representations into syllabic segments. However, when trained with an utterance-level cross-entropy objective, the model predicts speaker identity rather than linguistic content, thereby compromising the purity of syllabic tokens. To address this problem, we propose a speaker-disentangled syllabic tokenizer that regresses speaker-perturbed student representations toward clean teacher targets within fixed-length chunks. Experimental results demonstrate that our proposed method achieves state-of-the-art performance in syllable boundary detection and syllabic segment clustering. Moreover, a speech language model trained on our syllabic tokens achieves a 7\% relative improvement in syntactic and semantic understanding over the phone-level SpiRit-LM.
\end{abstract}


\maketitle

\section{INTRODUCTION}
\IEEEPARstart{S}{elf-supervised} speech representation learning has been shown to effectively extract phonetic content from raw speech~\cite{9585401,pmlr-v202-baevski23a}. This enables phonetic tokens to serve as pseudo-transcripts, thereby allowing language modeling directly on speech tokens~\cite{lakhotia-etal-2021}. As a result, speech language models (LMs) offer a unified framework for understanding and generating spoken language, and have emerged as a foundation for spoken dialogue modeling~\cite{zhang-etal-2023-speechgpt,10096250,defossez2024moshispeechtextfoundationmodel,zeng2025scaling}.

To transfer linguistic knowledge from text LMs to speech LMs, SpiRit-LM introduces word-level speech-text interleaving, where textually pretrained LMs are continually trained on sequences that alternate between phonetic and text tokens at word boundaries~\cite{10.1162/tacl_a_00728}. However, a fundamental challenge lies in the mismatch of token granularity between speech and text. Learned phonetic tokens typically occur at a high frame rate (12.5--50~Hz), whereas text is encoded using coarser subword tokens. This lower linguistic information density in speech tokens reduces computational efficiency and exacerbates the granularity mismatch, which can hinder speech-text alignment.

\begin{figure*}[!t]
\centering
\subfloat[HuBERT]{\includegraphics[width=0.23\textwidth]{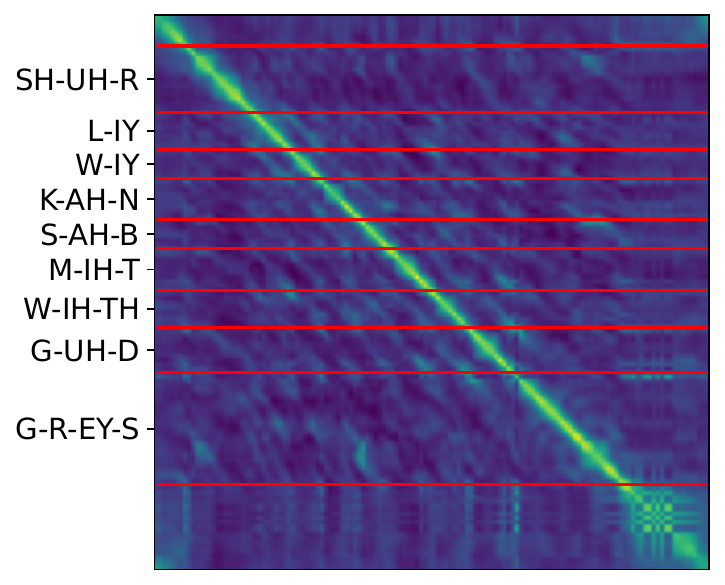}%
\label{fig:similarity_mat_hubert}}
\hfil
\subfloat[SD-HuBERT]{\includegraphics[width=0.23\textwidth]{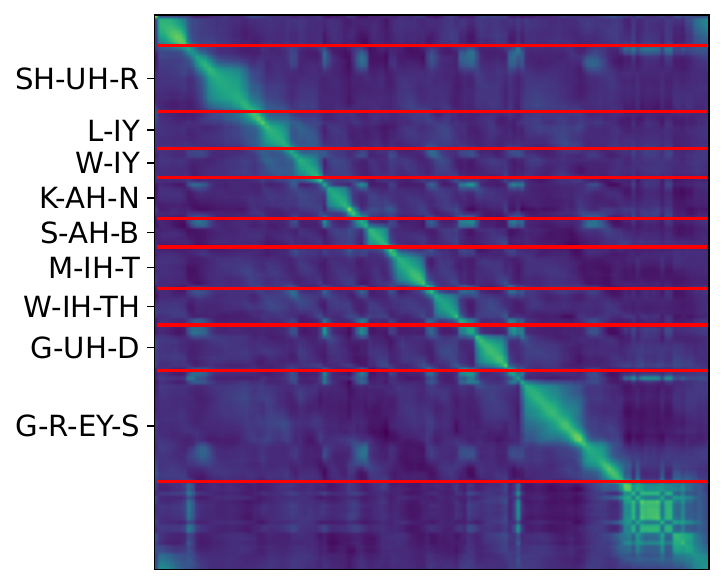}%
\label{fig:similarity_mat_sdhubert}}
\hfil
\subfloat[SylReg (ours)]{\includegraphics[width=0.23\textwidth]{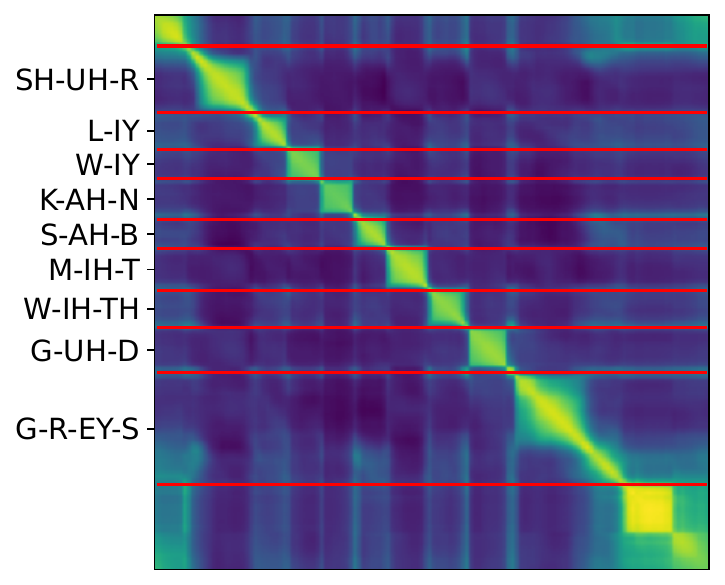}%
\label{fig:similarity_mat_ours}}
\caption{Self-similarity matrices of latent speech frame representations extracted from the $\ell$th Transformer layer, where $\ell=8$ for (a) HuBERT and (c) SylReg, and $\ell=9$ for (b) SD-HuBERT, following its official configuration. Red lines indicate ground-truth syllable boundaries. The transcript of the speech sample is ``Surely we can submit with good grace.''}
\label{fig:similarity_mat}
\end{figure*}

To mitigate this mismatch, recent approaches aim to discover linguistically meaningful, coarser syllabic tokens~\cite{peng23e_interspeech,10446062,Komatsu_Self-Supervised_Syllable_Discovery_2024,baade2024,cho2024sylber}. In particular, Cho \textit{et al.} proposed SD-HuBERT, a self-distillation framework for the pretrained HuBERT based on an utterance-level cross-entropy objective~\cite{Caron_2021_ICCV,10446062}. This approach implicitly organizes latent frame representations into syllabic segments in an intermediate Transformer layer~\cite{NIPS2017_3f5ee243}, as shown in Figure~\ref{fig:similarity_mat_sdhubert}~\cite{10446062}. Syllabic tokens are derived as cluster indices via a three-step tokenization procedure: 1) computing a self-similarity matrix of frame-level features, 2) segmenting the matrix to identify syllable boundaries, and 3) quantizing the segment-wise average features. Recent studies have shown that speech LMs built on these syllabic tokens outperform speech LMs trained on phone-level tokens in syntactic understanding, suggesting that they better capture linguistically abstract concepts~\cite{baade2024,cho2024sylber,kando25_interspeech}. Furthermore, syllabic tokens have demonstrated effectiveness in unsupervised speech recognition as well~\cite{wang2025unsupervisedspeechrecognitionsyllablelevel}.

However, we observed valid prototype collapse in SD-HuBERT, where only a small subset of the final softmax categories becomes active~\cite{10472570}. This collapse likely degrades syllabic features, as identical category signals are backpropagated to linguistically diverse utterances. Moreover, we found a moderate correlation between speaker identities and the final softmax categories~\cite{Komatsu_Self-Supervised_Syllable_Discovery_2024}. This suggests that SD-HuBERT tends to predict speaker identity, contaminating the purity of the syllabic tokens. We partially attribute this speaker-dominating problem to the utterance-level nature of the SD-HuBERT objective, as speaker characteristics tend to be stationary over an utterance and utterance-level representations have been shown to capture speaker information~\cite{niekerk21_interspeech}.

To address these two problems, we previously proposed a speaker-disentangled training objective based on frame-wise regression rather than utterance-level classification~\cite{Komatsu_Self-Supervised_Syllable_Discovery_2024}. In this framework, the student and its moving average teacher are trained to extract consistent frame-level representations from both the original waveform and its speaker-perturbed counterpart. By enforcing speaker-invariant consistency at the frame-level, the model is encouraged to focus on local, content-related structure. Moreover, regression objectives inherently avoid valid prototype collapse.

In this paper, we extend our previous work~\cite{Komatsu_Self-Supervised_Syllable_Discovery_2024} in two directions. First, we propose syllabic tokenization via chunk-wise regression (SylReg). Whereas our previous frame-wise regression disentangled speaker attributes, its locality does not sufficiently promote syllabic grouping. SylReg addresses this limitation by enforcing coherence over mid-level temporal chunks, enabling speaker-resilient yet syllabically structured representations. Experimental results show that SylReg outperforms state-of-the-art methods~\cite{cho2024sylber,baade2024} in syllable boundary detection and syllabic segment clustering.

Second, we introduce SylReg-LM, a speech LM trained on interleaved syllabic and text tokens. SylReg-LM achieves a 7\% average relative improvement on syntactic and semantic understanding over SpiRit-LM~\cite{10.1162/tacl_a_00728}, a phone-level-interleaved speech LM, demonstrating its efficacy in capturing high-level linguistic abstractions. In addition, we successfully train a token-to-speech synthesizer that matches the TWIST synthesizer~\cite{NEURIPS2023_c859b99b} in terms of character and word error rates, whereas using a 2.3$\times$ lower token bitrate. Our code, models, and generated speech samples are publicly available.~\footnote{\url{https://github.com/ryota-komatsu/speaker_disentangled_hubert}}

We summarize our contributions as follows.
\begin{enumerate}
\item We identify a bias toward speaker identity in SD-HuBERT, contaminating the syllabic token purity.
\item We propose an unsupervised framework designed to learn speaker-disentangled syllabic representations by optimizing a chunk-wise regression objective.
\item We demonstrate that our proposed method achieves state-of-the-art performance in syllabic tokenization.
\item We introduce an interleaved syllable-text LM that improves high-level linguistic understanding compared to the phone-level token-based SpiRit-LM.
\item We demonstrate that our syllabic tokens enable intelligible speech synthesis with high coding efficiency.
\end{enumerate}

\section{Related Work}\label{sec:related_work}

\subsection{Trade-off between phonetic and acoustic tokens}
Speech information can be represented using two types of tokens with distinct properties. Phonetic tokens are obtained by quantizing latent representations extracted from self-supervised speech encoders~\cite{9585401,NEURIPS2023_c859b99b} or automatic speech recognition models~\cite{zeng2025scaling}. Owing to their strong alignment with underlying linguistic content, speech LMs built on these phonetic tokens enable intelligible speech generation~\cite{lakhotia-etal-2021}. However, fine-grained acoustic details are largely marginalized. In contrast, acoustic tokens are produced by neural audio codecs trained to faithfully reconstruct input waveforms~\cite{9625818}. Although acoustic tokens can encode general audio, including environmental sounds and music, they are weakly aligned with textual content~\cite{zhang2024speechtokenizer}. This misalignment hinders lexical, syntactic, and semantic understanding in speech LMs~\cite{mousavi2025discrete}. In this work, we focus on learning phonetic tokens for linguistic content modeling. To combine the complementary strengths of both token types, phonetic tokens can be integrated into neural audio codecs via semantic distillation~\cite{zhang2024speechtokenizer,defossez2024moshispeechtextfoundationmodel}.

\subsection{Learning coarse phonetic tokens for speech LMs}
An orthogonal line of research explores learning coarser subword- or syllable-level phonetic tokens. Some training-free approaches apply subword tokenization or deduplication to phonetic tokens~\cite{10446063,visser25_interspeech}. Although simple, these methods still operate at relatively high token frame rates of around 20~Hz. Another direction inserts speech adapters into the LM front-end to aggregate frame-level tokens into coarser representations~\cite{lu2025latentspeechtexttransformer,cuervo2025latefusionmultilevelfission}. However, this often introduces additional complexity in training speech LMs, such as the need for curriculum learning~\cite{lu2025latentspeechtexttransformer} or the risk of model collapse~\cite{cuervo2025latefusionmultilevelfission}. In contrast, syllabic tokenizers shift temporal aggregation to the tokenization stage, allowing LMs to remain architecturally unchanged~\cite{baade2024,cho2024sylber}. Within this category, Sylber 2.0 adopts speaker-disentangled frame-wise regression~\cite{Komatsu_Self-Supervised_Syllable_Discovery_2024} and extends it to multilingual acoustic features for language-universal, expressive speech synthesis~\cite{cho2026sylber20universalsyllable}. In this work, we advance this framework by introducing a chunk-wise regression, which enables more accurate syllabic segmentation in English. Moreover, a concurrent work, ZeroSyl, introduces distillation-free syllabic tokenization by directly utilizing an off-the-shelf speech encoder~\cite{visser2026zerosylsimplezeroresourcesyllable}. While ZeroSyl focuses on LMs trained exclusively on syllabic tokens, our interleaved syllable-text LMs enhance semantic modeling through knowledge transfer from text.

\section{Self-Supervised Syllabic Tokenization}\label{sec:syllable_discovery}

\begin{figure*}[t!]
\centering
\subfloat[Speaker-disentangled chunk-wise regression]{\includegraphics[width=0.39\textwidth]{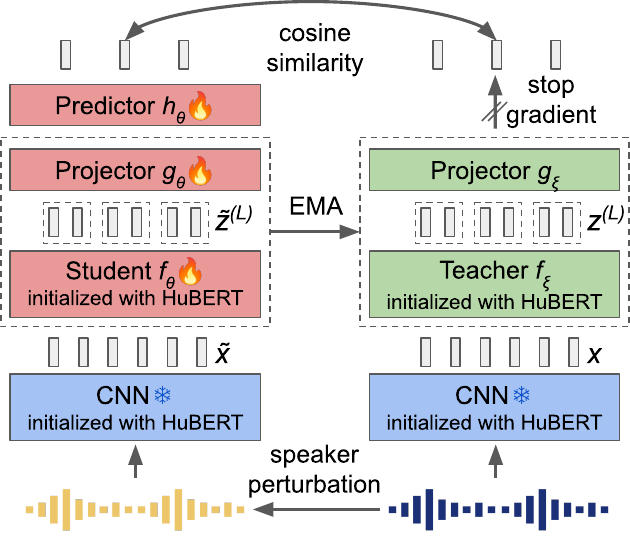}%
\label{fig:method1}}
\hfil
\subfloat[Speech language modeling]{\includegraphics[width=0.39\textwidth]{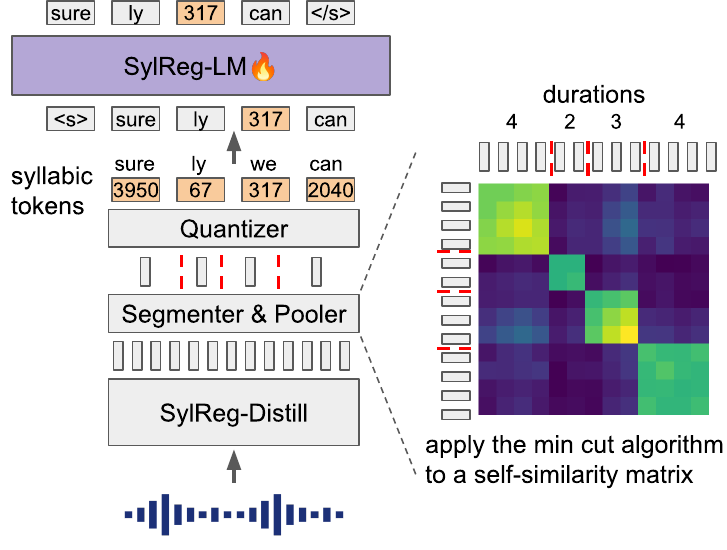}%
\label{fig:method2}}
\caption{Our proposed method. The left (a) and right (b) figures illustrate the syllabic tokenization and language modeling phases, respectively.}
\label{fig:method}
\end{figure*}

\subsection{Baseline method: SD-HuBERT}\label{sec:baseline}
SD-HuBERT~\cite{10446062} finetunes the pretrained HuBERT using DINO~\cite{Caron_2021_ICCV}, a self-distillation framework that has demonstrated emergent image segmentation capabilities. Similar to BERT, a learnable classification token \texttt{[CLS]} is prepended to the input speech frame sequence $x=[x_1,\dots,x_T]$ and aggregated via self-attention layers to obtain an utterance-level embedding $z_\texttt{[CLS]}^{(L)}$ from the last layer $L$~\cite{devlin-etal-2019-bert}. A classification head computes logits as the dot products between $z_\texttt{[CLS]}^{(L)}$ and a set of learnable prototypes $\phi$, each corresponding to a softmax category. A pseudo-category $c$ is then predicted from the resulting categorical distribution $p_\phi(c\mid z_\texttt{[CLS]}^{(L)})$. The student and its moving average teacher are optimized to minimize an utterance-level cross-entropy objective. Through distillation, syllabic organization emerges in the latent speech frame representations $z^{(\ell)}=[z_1^{(\ell)},\dots,z_T^{(\ell)}]$ extracted from the $\ell(=9)$th Transformer layer of the student. As shown in Figure~\ref{fig:similarity_mat_sdhubert}, constructing a self-similarity matrix $z^{(\ell)}{z^{(\ell)}}^\top\in\mathbb{R}^{T\times T}$ reveals block structures whose boundaries roughly match the ground truth. To find syllabic segments from this matrix, the minimum cut algorithm~\cite{malioutov-barzilay-2006-minimum,peng23e_interspeech} is used, with an algorithmic improvement for computational efficiency.

\subsection{Speaker-dominating problem in SD-HuBERT}\label{sec:analysis}
In~\cite{10446062}, experimental results on sentence discriminability suggested that paralinguistic or nonlinguistic information might dominate the aggregated \texttt{[CLS]} representation. To investigate this question, we computed the speaker-normalized mutual information $I(X; Y) / H(X) = 1 - H(X\mid Y) / H(X)$ between the speaker ID $X$ and the predicted category $Y=\argmax p_\phi(c\mid z_\texttt{[CLS]}^{(L)})$. This metric quantifies the relative reduction in entropy (uncertainty) about speaker identities after observing the predicted categories~\cite{9585401}. We observed a large value of 0.61 on the LibriSpeech~\cite{7178964} test set, indicating that the model tends to discriminate speaker identity rather than linguistic content. This finding is consistent with Niekerk \textit{et al.}, who showed that utterance-wise mean frame representations capture speaker identity, as such characteristics remain relatively stationary over an utterance~\cite{niekerk21_interspeech}. Indeed, the self-attended \texttt{[CLS]} representation can be interpreted as a weighted global average of frame representations.

\subsection{SylReg: Speaker-disentangled chunk-wise regression}\label{sec:proposed}
Motivated by the speaker-dominating problem in SD-HuBERT, we propose a speaker-disentangled objective that emphasizes linguistic content by matching speaker-invariant representations between the original speech and its speaker-perturbed counterpart within fixed-length chunks. Figure~\ref{fig:method1} illustrates our proposed SylReg, which follows a BYOL-style framework~\cite{NEURIPS2020_f3ada80d}. The model consists of two branches: a student and a teacher. The student comprises a convolutional neural network (CNN) encoder, a Transformer encoder $f_\theta$, a projector $g_\theta$, and a predictor $h_\theta$. The teacher, parameterized by $\xi$, shares the same backbone architecture but omits the predictor. We initialize the CNN and Transformer encoders using the pretrained HuBERT Base. Following~\cite{baade2024,cho2024sylber}, we prune the last three Transformer layers, resulting in a nine-layer Transformer. The last three pretrained layers have been shown to hinder models from learning linguistically coarse syllabic representations~\cite{10446062}.

The student is trained to regress the teacher's projected hidden states. Unlike BYOL, which performs global-level regression, our objective operates on fixed-length chunks to encourage mid-level temporal grouping, which empirically aligns with syllables. Prior to projection, average pooling with a chunk size of $C$ is applied to aggregate hidden states from the final Transformer layer $L$:
\begin{align*}
[\tilde{z}_1^{(L)},\tilde{z}_2^{(L)},\dots,\tilde{z}_{T/C}^{(L)}]=&f_\theta(\tilde{x}),\\
[z_1^{(L)},z_2^{(L)},\dots,z_{T/C}^{(L)}]=&f_\xi(x),
\end{align*}
where $x$ and $\tilde{x}$ denote speech frame features of the original and speaker-perturbed speech, respectively. This reduces the sequence length by a factor of $C$. We then minimize the mean squared error (MSE) between $\ell_2$-normalized teacher and student outputs:
\begin{align*}
\mathcal{L}_\mathrm{SylReg} =& 
\sum_{t=1}^{T/C}
\left\lVert
\frac{(h_\theta\circ g_\theta)(\tilde{z}_t^{(L)})}{\lVert(h_\theta\circ g_\theta)(\tilde{z}_t^{(L)})\rVert}
-
\frac{g_\xi(z_t^{(L)})}{\lVert g_\xi(z_t^{(L)})\rVert}
\right\rVert^2,
\end{align*}
where $z_t^{(L)}$ and $\tilde{z}_t^{(L)}$ denote the $t$th chunk-level representations of the original and perturbed speech, respectively. During training, gradients are stopped on the teacher outputs so that only the student is updated via backpropagation. For efficient training, we freeze the CNN in both the teacher and student. The remaining teacher parameters are updated using an exponential moving average (EMA) of the student.

To disentangle speaker-specific characteristics while preserving linguistic content, we adopt the speaker perturbation method proposed in~\cite{NEURIPS2021_87682805} with a modification. The original algorithm applies random formant shifts and pitch perturbations. However, we observe that unconstrained perturbations, e.g., doubling the pitch of a female voice, often produce unnatural speech. To ensure naturalness, we restrict formant shifts and pitch perturbations to male-to-female and female-to-male conversions only. We estimate the speaker's gender based on the average pitch of each utterance. If the mean pitch exceeds a predefined threshold, we apply a female-to-male conversion; otherwise, we apply a male-to-female conversion. The original speech and its perturbed counterpart are fed into the teacher and student, respectively.

\subsection{Self-segmentation distillation}\label{sec:self-segment}
As shown in Figure~\ref{fig:similarity_mat_ours}, SylReg induces a block-diagonal syllabic structure in the student's 8th Transformer layer. We distill this emergent structure into the pretrained data2vec 2.0~\cite{pmlr-v202-baevski23a} via self-segmentation distillation (SylBoost)~\cite{baade2024,cho2024sylber} and refer to the resulting model as SylReg-Distill. We adopt data2vec 2.0 because its representations enable better syllabic segment clustering than those of HuBERT~\cite{baade2024}.

Self-segmentation distillation proceeds in multiple stages. In the first stage, the SylReg student is used to segment each utterance into pseudo-syllable boundaries using the algorithm described in Section~\ref{sec:gslm}-\ref{sec:quantizer}. We then initialize a new teacher-student pair from data2vec 2.0 Base and prune the 12th Transformer layer~\cite{baade2024}. The teacher's frame representations are averaged within each pseudo-syllable segment to form regression targets. The student's frame representation at time step $t$ is regressed toward the teacher's syllabic embedding for the segment $S_t$ that contains frame $t$:
\begin{align*}
\mathcal{L}_\mathrm{Self-segment}=\frac{1}{T}\sum_{t=1}^T\left\lVert \tilde{z}_t^{(L)}-\frac{1}{\lvert S_t\rvert}\sum_{s\in S_t}z_s^{(L)}\right\rVert^2,
\end{align*}
where $z_t^{(L)}$ and $\tilde{z}_t^{(L)}$ denote the $t$th speech frame representations of the teacher and student, respectively. Speaker perturbation is applied to the student inputs. The teacher is iteratively updated by copying the student parameters after each stage. In subsequent stages, the updated teacher is used both for extracting and segmenting regression targets, enabling iterative self-refinement of the syllabic structure.

\section{Generative Spoken Language Modeling}~\label{sec:gslm}
The generative spoken language modeling (GSLM) pipeline consists of three modules: a speech tokenizer, a speech LM, and a token-to-speech synthesizer, which we describe below.

\subsection{Syllable segmentation}~\label{sec:quantizer}
To detect syllable boundaries and quantize syllabic representations, we use the syllabic tokenization algorithm proposed in~\cite{peng23e_interspeech}, with a PyTorch-based implementation for improved efficiency. As depicted in Figure~\ref{fig:method2}, the procedure comprises three steps: 1) minimum cut segmentation, 2) segment-wise average pooling, and 3) two-step clustering. Given the student's $\ell$th Transformer layer outputs $z^{(\ell)}=[z_1^{(\ell)},\dots,z_T^{(\ell)}]$, we first compute a frame-level self-similarity matrix $z^{(\ell)}{z^{(\ell)}}^\top\in\mathbb{R}^{T\times T}$. We then apply the minimum cut algorithm to obtain $M$ syllable segments, where $M$ is predefined as $\lceil T\cdot F/50\rceil$ based on the HuBERT frame rate of 50~Hz and the upper-bound syllabic token frame rate $F$. For efficient computation, we implement this algorithm as an $O(M)$-complexity dynamic programming loop. To account for fast speaking rates, we first oversegment speech frames and then merge adjacent segments whose mean features have a cosine similarity greater than a threshold $\tau$. After detecting syllable boundaries, we average the features within each segment. Finally, we apply $K$-Means clustering to obtain centroids and then perform agglomerative clustering over these centroids to assign syllabic tokens to each segment.

\subsection{Interleaved syllable-text language modeling}\label{sec:speech_lm}
Following SpiRit-LM~\cite{10.1162/tacl_a_00728}, we expand the original textual vocabulary with syllabic tokens and continually train a textually pretrained LM on sequences that interleave text and syllabic tokens. Given a transcribed utterance, we first obtain word-level timestamps using a forced aligner. Each word is then aligned with its syllabic tokens by matching these timestamps with syllable segmentation results. To construct interleaved syllable-text data, we randomly sample either syllabic tokens or text from each chunk. For example, an aligned utterance [\{``text'': ``Surely'', ``speech'': [3950, 67], ``time'': [0.0, 0.5]\}, \{``text'': ``we'', ``speech'': [317], ``time'': [0.5, 0.7]\}, \{``text'': ``can'', ``speech'': [2040], ``time'': [0.7, 1.0]\}] can be interleaved as ``Surely\textlangle 317\textrangle can.''

\subsection{Token-to-speech synthesis}\label{sec:u2s}
We adopt a conditional flow-matching (CFM)-based Diffusion Transformer (DiT) with a BigVGAN-v2 vocoder conditioned on log mel-spectrograms~\cite{NEURIPS2023_2d8911db,10377858,lee2023bigvgan}. CFM-based DiTs are well suited for speech LMs, as they enable fast generation of acoustic features from speech tokens with fewer than ten non-autoregressive denoising steps. To generate acoustic features in parallel, the length regulator first takes syllabic representations from the input embedding layer, then repeats them according to their predicted durations (i.e., the number of mel frames each syllable spans), and finally feeds the expanded sequence into the DiT layers~\cite{NEURIPS2019_f63f65b5}. The length regulator is jointly trained with DiT to regress token durations derived from syllable segmentation. We initialize and freeze the input embeddings using the pretrained syllabic clustering centroids for efficient knowledge transfer.

\section{Experimental Setup}\label{sec:exp_setup}

\subsection{Datasets}
We train SylReg on Libri-Light~\cite{9052942}, a 55k hours of audio book corpus, and then distill syllable segments using the LibriSpeech train-clean-100 subset. Quantizers are trained on the LibriSpeech train set and evaluated on its test set using syllable alignments~\cite{10446062}. For language modeling, we use 129k hours of English speech corpora, including LibriSpeech, Libriheavy~\cite{10447120}, Emilia-Large~\cite{11175027}, People's Speech (\texttt{clean} and \texttt{clean\_sa} subsets)~\cite{galvez2021the}, VoxPopuli (transcribed subset)~\cite{wang-etal-2021-voxpopuli}, and synthetic speech generated from TinyStories~\cite{eldan2023tinystoriessmalllanguagemodels} and Cosmopedia v2~\cite{allal2024SmolLM} using Kokoro text-to-speech with the \texttt{af\_heart} voice~\cite{hexgrad_2025}. We retain utterances from Emilia-Large with a mean opinion score (MOS) above 3.45 and a duration exceeding 10 seconds, and further filter out utterances with transcription errors or code-switching, following~\cite{chen-etal-2025-f5}. To mitigate catastrophic forgetting on text, we mix Cosmopedia v2. Finally, we pretrain a token-to-speech synthesizer on LibriTTS-R~\cite{koizumi23_interspeech} and finetune it on the female subset of Hi-Fi-CAPTAIN~\cite{hi-fi-captain} for single-speaker synthesis.

\subsection{Implementation details}
\textbf{Speech encoder}
We set the default chunk size $C$ to 100 frames. Both the projector and the predictor are 2-layer multilayer perceptrons, each comprising a linear layer with an output size of 2048 followed by a batch normalization~\cite{pmlr-v37-ioffe15}, the GELU~\cite{hendrycks2023gaussian} activation function, and another linear layer with an output size of 256. We train SylReg for 10k steps using AdamW~\cite{loshchilov2018decoupled} with a weight decay of 0.01, a gradient norm clipping of 1e-3, an EMA decay of 0.999, and a batch size of 1024. The learning rate (LR) is fixed at 1e-4 with linear warmup over the first 100 steps. We update only the randomly initialized projector and predictor during the first 2k steps. After chunk-wise regression, self-segmentation distillation is performed in five stages (200 steps followed by four stages of 50 steps), using the same optimizer settings as SylReg. For speaker perturbation, we set the parameters (formant shift ratio, new pitch median, pitch range factor) to (1.1, 300, 1.2) and (1/1.1, 100, 1/1.2) for male-to-female and female-to-male conversions, respectively, using a mean pitch threshold of 155~Hz to classify the conversion types. We trained models on four NVIDIA A6000 GPUs.

\noindent\textbf{Tokenizer}
Following~\cite{baade2024}, we set the minimum and maximum segment durations to 3 and 35 frames, respectively. The upper-bound syllabic token frame rate $F$ and the merge threshold for SylReg-Distill are fixed at 6.67~Hz and 0.95, respectively. For SylReg, we tune the merge threshold $\tau_\mathrm{SylReg}=0.7$ to achieve a token frame rate of approximately 6.25~Hz, which has been shown to perform best in speech language modeling~\cite{baade2024}. We use layer $\ell=8$ in SylReg and $\ell=11$ in SylReg-Distill. For HuBERT, we follow SylReg and use $\ell=8$. For other models, we adopt their default configurations: SD-HuBERT and Sylber employ the 9th layer, whereas SylBoost uses the 11th layer. In SylReg-Distill, we configure $K$-Means with 24576 centroids and agglomerative clustering with 8192 clusters. For comparison with prior work~\cite{10446062,cho2024sylber}, we additionally evaluate SylReg using $K=16384$ and 4096 agglomerative clusters. We train $K$-Means for 50 iterations with 5 random initializations.

\begin{table*}[t]
\centering
\caption{Syllable segmentation scores, syllabic token quality, and token edit distance (TED) on the LibriSpeech test set. Rows are grouped by vocabulary size.}
\label{tab:segmentation}
\begin{tabular}{lccccccccccccc}\hline
\multirow{2}{*}{Model}&\multirow{2}{*}{Initialization}&\multirow{2}{*}{w/ distill}&Vocab&Token&\multicolumn{4}{c}{Syllable segmentation (\%)\textuparrow}&&\multicolumn{3}{c}{Token purity (\%)\textuparrow}&\multirow{2}{*}{TED (\%)\textdownarrow}\\\cline{6-9}\cline{11-13}
&&&size&frame rate&Pr&Re&F1&R&&SP&CP&SNMI&\\\hline
Ground truth&&&4973&4.38&&&&&&&\\\hline
HuBERT~\cite{9585401}&                      &&4096&6.71&47.9&75.9&58.7&39.1&&61.8&33.9&80.9&16.6\\
SD-HuBERT~\cite{10446062}&HuBERT            &&4096&4.67&64.3&71.0&67.5&70.7&&54.1&\textbf{46.2}&73.4&19.4\\
\rowcolor{cyan!10}SylReg (ours)&HuBERT  &&4096&6.31&60.3&\textbf{89.8}&72.2&54.1&&\textbf{70.5}&42.5&\textbf{86.0}&\textbf{9.91}\\
Sylber~\cite{cho2024sylber}&SD-HuBERT       &\checkmark&4096&3.76&\textbf{76.6}&68.3&72.2&\textbf{75.9}&&64.0&43.9&81.4&14.1\\
\hline
SylBoost 6.25Hz~\cite{baade2024}&data2vec 2.0&\checkmark&8192&5.86&64.1&88.7&74.4&62.4&&76.6&33.9&90.2&14.4\\
\rowcolor{cyan!10}SylReg-Distill (ours)&data2vec 2.0&\checkmark&8192&5.82&64.5&88.7&\textbf{74.7}&63.1&&\textbf{79.5}&\textbf{34.9}&\textbf{91.5}&\textbf{7.66}\\
\hline
\end{tabular}
\end{table*}

\noindent\textbf{Speech language model}
We continually train Qwen2.5 7B~\cite{qwen2025qwen25technicalreport} for 15k steps using AdamW (weight decay 0.01, $\beta_1=0.9$, $\beta_2=0.95$), a gradient norm clipping of 0.5, and a batch size of 2.1 million tokens. We use a trapezoidal LR scheduler with peak/minimum LRs of 3e-4/3e-5, 100 warmup steps, and 5k decay steps. We also train 85M LMs for 50k steps with peak/minimum LRs of 5e-4/5e-5 and a batch size of 320k tokens. For interleaving, during the first 5k steps, we update only the randomly initialized syllabic token embeddings on speech-only and interleaved data. We align speech segments with their transcripts using the NeMo Forced Aligner~\cite{rastorgueva23_interspeech}. For interleaving, we sample syllabic tokens with a probability of approximately 0.3 and otherwise select text tokens. The training data comprise speech-only, text-only, and interleaved examples in a roughly 3:3:4 ratio. The training took 39 hours on 32 H100 GPUs.

\noindent\textbf{Token-to-speech synthesizer}
We extract 80-bin log mel-spectrograms using an STFT window size of 400 and a hop size of 320, and standardize them following~\cite{NEURIPS2023_2d8911db}. DiT consists of feed-forward Transformer blocks~\cite{NEURIPS2019_f63f65b5}, comprising a 2-layer encoder with hidden/intermediate sizes of 768/1536 and a 4-layer decoder with 512/1024. We use QK-Norm to stabilize training~\cite{henry-etal-2020-query}. The length regulator consists of a single convolution layer with a kernel size of 3. We pretrain/finetune DiT for 200k/50 steps with a gradient norm clipping of 0.1, batch sizes of 400/14k sentences, LRs of 1e-3/1e-4, and 1k linear warmup steps. During training, we drop the entire syllabic token sequence with a probability of 0.2 for classifier-free guidance (CFG)~\cite{ho2021classifierfree}. At inference time, we set the step size in the Euler method to 0.1 and the strength of CFG to 0.7.
For BigVGAN, we follow the original setup except for the following modifications. We use the BigVGAN-base architecture with upsampling kernel sizes of [10, 9, 8, 4, 4] and strides of [5, 4, 4, 2, 2] to synthesize 16~kHz waveforms. We train BigVGAN for 1M steps with a gradient clipping of 100 and a batch size of 20-second speech segments. We used two A6000 GPUs.

\section{Evaluation}\label{sec:results}
\subsection{Syllable segmentation and token quality}
\noindent\textbf{Metrics}
Following~\cite{10446062}, we measure precision (Pr), recall (Re), F1, and R-value (R)~\cite{rasanen09b_interspeech} of syllable boundaries with a tolerance of 50~ms. The R-value is a comprehensive metric that balances the trade-off between recall and over-segmentation. Additionally, we evaluate the quality of the syllabic tokens using syllable purity (SP), cluster purity (CP), and syllable-normalized mutual information (SNMI)~\cite{9585401}:
\begin{align*}
\mathrm{SP}=&\mathbb{E}_{p(u)}[p(\argmax_s p(s,u)\mid u)],\\
\mathrm{CP}=&\mathbb{E}_{p(s)}[p(\argmax_u p(s,u)\mid s)],\\
\mathrm{SNMI}=&I(S;U)/H(S)=1-H(S\mid U)/H(S),
\end{align*}
where $p(s,u)$ denotes the joint distribution of the ground-truth syllable $S$ and the syllabic token $U$. SNMI quantifies the fraction of syllable entropy explained by the tokens. We align the reference and predicted syllables using maximum weight matching on the temporal intersection-over-union matrix of their segments. Token quality is fairly comparable across models with the same vocabulary size~\cite{9585401}, as increasing the vocabulary size improves SP by reducing syllable mixing within clusters, but decreases CP by more frequently splitting the same syllable across multiple clusters. To assess speaker dominance directly at the syllabic tokens, we measure the token edit distance (TED)~\cite{gat-etal-2023-augmentation} between the original syllabic token sequence $\bm{u}$ and its speaker-perturbed counterpart $\bm{\tilde{u}}$ (See Section~\ref{sec:syllable_discovery}-\ref{sec:proposed} for perturbation.). Formally, TED is defined as $\editdist(\bm{u}, \bm{\tilde{u}})/\len(\bm{u})$, which quantifies the discrepancy in syllabic tokens caused by speaking variations.

\noindent\textbf{Results}
Table~\ref{tab:segmentation} summarizes the segmentation scores and clustering quality on the LibriSpeech test set. SylReg consistently outperforms HuBERT in both syllable segmentation and token purity. This suggests that naive segmental pooling of phone-level representations degrades syllabic representations by averaging heterogeneous frames. Sylber achieves the highest precision, as it removes silence and thus has the lowest token frame rate. SylReg matches Sylber on segmentation F1 and outperforms it on SP by 10\%, without relying on self-segmentation distillation. When SylReg is distilled into data2vec 2.0, SylReg-Distill matches or exceeds SylBoost across all syllable segmentation and token purity metrics. SD-HuBERT exhibits the highest TED, suggesting that its syllabic tokens are sensitive to speaking variations. Since LibriSpeech is a multi-speaker dataset, high token purity reflects stronger faithfulness to linguistic content.

\begin{table}[t]
\centering
\caption{Ablation study on the LibriSpeech test set.}
\label{tab:ablation}
\resizebox{\linewidth}{!}{
\begin{tabular}{lccccccccc}\hline
\multirow{2}{*}{Model}&\multicolumn{4}{c}{Syllable segmentation (\%)\textuparrow}&&\multicolumn{3}{c}{Token purity (\%)\textuparrow}&\multirow{2}{*}{TED (\%)\textdownarrow}\\\cline{2-5}\cline{7-9}
&Pr&Re&F1&R&&SP&CP&SNMI&\\\hline
HuBERT&47.9&75.9&58.7&39.1&&61.8&33.9&80.9&16.6\\
SylReg&\textbf{60.3}&\textbf{89.8}&\textbf{72.2}&\textbf{54.1}&&\textbf{70.5}&\textbf{42.5}&\textbf{86.0}&\textbf{9.91}\\
-- BYOL + DINO&56.3&88.8&68.9&46.3&&69.0&40.3&85.2&11.9\\
\quad-- Speaker perturb&56.3&88.7&68.8&46.3&&68.7&39.9&85.1&13.9\\
+ Random perturb&58.8&87.1&70.2&53.4&&69.7&41.7&85.5&12.2\\
\hline
\end{tabular}
}
\end{table}

\subsection{Ablation study}
Table~\ref{tab:ablation} summarizes the effect of architectural and training variations. We first examine the impact of the model architecture by replacing BYOL with DINO and using a cross-entropy loss instead of MSE. Following SD-HuBERT, we configure DINO with 2048 classes, a student temperature of 0.2, a teacher temperature of 0.05, and a gradient clipping value of 0.5. Across all metrics, the DINO variant exhibits a uniform degradation, in line with findings in image segmentation~\cite{Caron_2021_ICCV}. We further observe that only 32\% of the categories in the final softmax function become active on the LibriSpeech test set, consistent with the general trend reported in~\cite{10472570}. This collapse in category utilization likely degrades syllabic representations, as identical category signals are backpropagated to linguistically diverse utterances. Removing speaker perturbation from the DINO variant increases TED and thus reduces the token purity, as syllables with speaking variations are prone to being assigned to different clusters. Therefore, while the DINO variant naturally reduces TED from 16.6 to 13.9, this inherent property alone may not be sufficient for high-purity syllabic tokenization. Replacing our speaker perturbation with a random perturbation~\cite{NEURIPS2021_87682805} leads to consistent performance degradation, demonstrating the efficacy of our perturbation design.

\begin{table}[t]
\centering
\caption{Effect of chunk size on syllable segmentation scores and syllabic token quality on the LibriSpeech development set.}
\label{tab:chunk}
\resizebox{\linewidth}{!}{
\begin{tabular}{lcccccccc}\hline
Chunk size&\multicolumn{4}{c}{Syllable segmentation (\%)\textuparrow}&&\multicolumn{3}{c}{Token purity (\%)\textuparrow}\\\cline{2-5}\cline{7-9}
$C$&Pr&Re&F1&R&&SP&CP&SNMI\\\hline
1&53.8&83.9&65.6&45.5&&69.7&41.6&85.8\\
20&59.1&89.1&71.1&52.2&&\textbf{70.5}&42.7&\textbf{86.2}\\
35&60.2&89.3&72.0&54.4&&70.3&43.0&86.1\\
50&60.7&89.4&72.3&55.3&&70.1&42.1&86.1\\
100&\textbf{60.9}&89.6&\textbf{72.5}&\textbf{55.6}&&70.3&\textbf{43.1}&86.0\\
$T(\le250)$&60.8&\textbf{89.7}&\textbf{72.5}&55.2&&69.9&42.6&85.8\\
\hline
\texttt{[CLS]}&47.3&24.2&32.1&45.4&&35.6&19.6&66.8\\\hline
\end{tabular}
}
\end{table}

\subsection{Effect of chunk size on SylReg}\label{sec:chunk-analysis}
Table~\ref{tab:chunk} shows the effect of chunk size on SylReg. The extreme case of $C=1$ reduces to frame-wise regression, whereas $C=T$ is equivalent to global average pooling, where $T$ is bounded by the maximum frame length of 250~\cite{10446062}. The trends differ between segmentation and clustering. Segmentation performance consistently improves as the chunk size increases, and saturates at a chunk size of 100. In contrast, SP and SNMI peak at a shorter chunk size of 20, which roughly corresponds to the 90th percentile of syllable durations in LibriSpeech. In addition, both frame-wise and global objectives result in suboptimal token purity. Therefore, optimizing chunk-wise regression at an intermediate temporal scale is crucial for learning syllabic representations that balance segmentation accuracy and token purity. We also observe that \texttt{[CLS]}-based aggregation yields poor results compared to global average pooling, suggesting that SylReg works better with average pooling.

\begin{figure}[t!]
\centering
\includegraphics[width=0.66\linewidth]{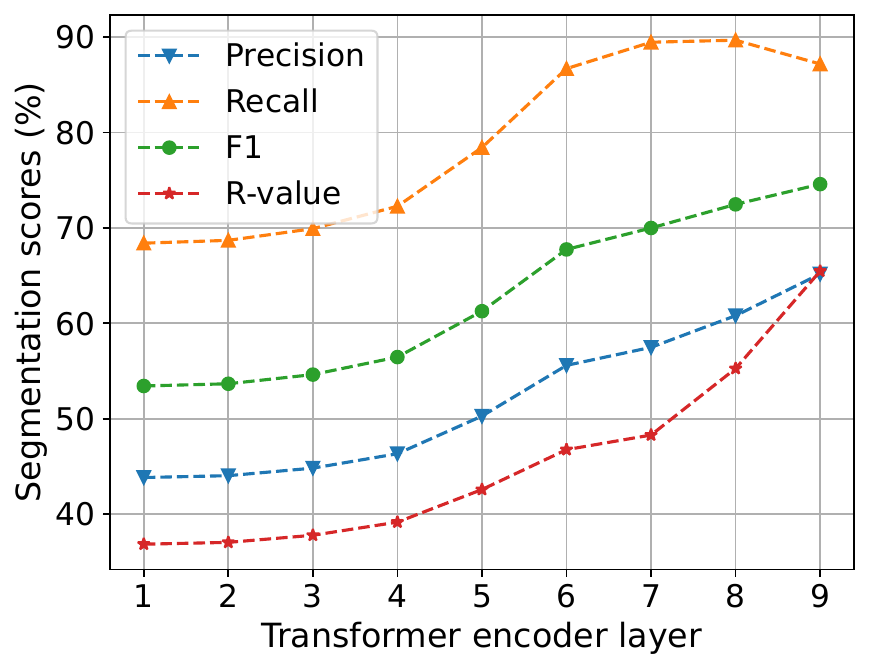}
\caption{Layer-wise syllable segmentation scores of SylReg on the LibriSpeech development set.\label{fig:layer-wise}}
\end{figure}

\subsection{Layer-wise analysis of segmentation performances}\label{sec:layer-wise-analysis}
Figure~\ref{fig:layer-wise} shows layer-wise syllable segmentation scores of SylReg on the LibriSpeech development set. The horizontal axis indicates the student layer used for feature extraction. In the early layers, all scores gradually improve with increasing depth. Recall peaks at the 8th Transformer layer and decreases at the 9th layer. Although the 9th layer yields the highest precision, this precision can be improved in the subsequent self-segmentation distillation stage. We therefore use the 8th layer for syllabic tokenization.

\subsection{Learning dynamics of SylReg}
\begin{figure}[t!]
\centering
\subfloat[Learning curve.]{\includegraphics[width=0.32\textwidth]{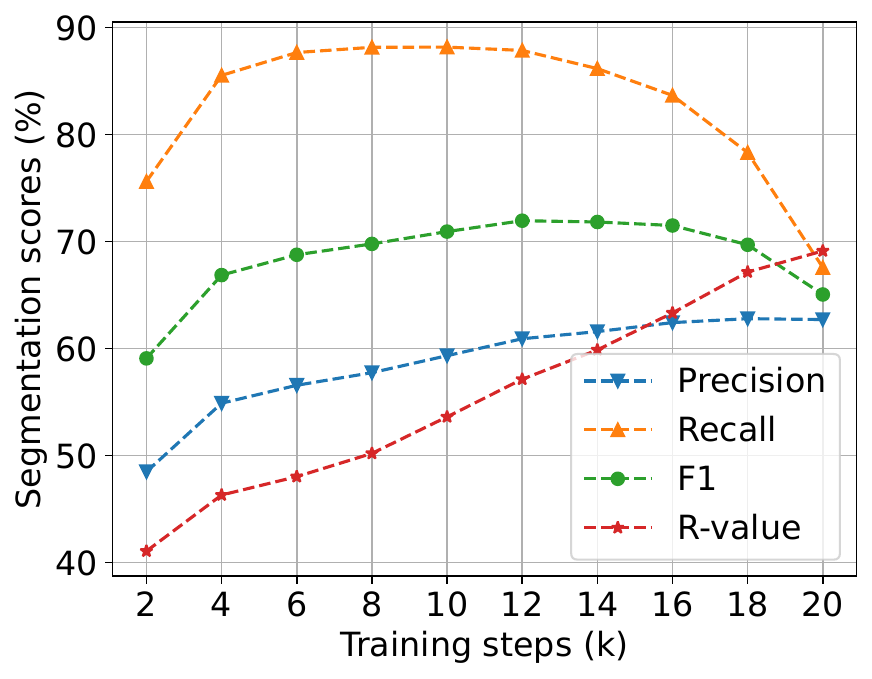}%
\label{fig:7}}
\hfil
\subfloat[Similarity matrices.]{\includegraphics[width=0.16\textwidth]{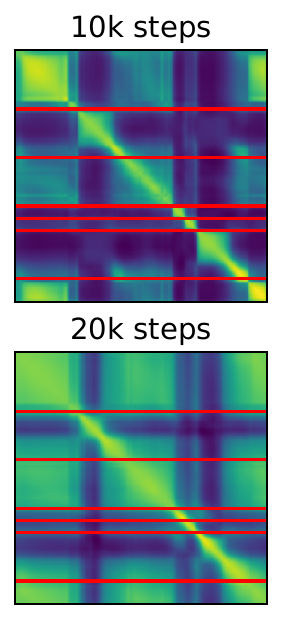}%
\label{fig:8}}
\caption{Learning dynamics of SylReg on the LibriSpeech development set. In Figure~\ref{fig:8}, red lines indicate ground-truth syllable boundaries.}
\end{figure}

Figure~\ref{fig:7} plots the learning curve of SylReg on the LibriSpeech development set. Recall peaks at 10k steps, and the overall F1 begins to decline after 12k steps. As depicted in Figure~\ref{fig:8}, the boundaries between adjacent syllables become blurred, and the similarity matrix shifts toward a more uniform distribution. Consequently, 10k training steps represent a reasonable choice to prevent representation collapse.

\begin{table*}[t]
\centering
\caption{Pretraining results of speech LMs on linguistic understanding and generation. ${}^\dagger$Understanding results are reported in~\cite{10.1162/tacl_a_00728}.
}
\label{tab:gslm}
\resizebox{\linewidth}{!}{
\begin{tabular}{lcccccccccccc}\hline
\multirow{2}{*}{Model}&\multirow{2}{*}{Encoder}&\multirow{2}{*}{w/ text}&Model&Data&Compute&Lexicon&Syntax&\multicolumn{2}{c}{Semantics}&&\multicolumn{2}{c}{Generation}\\\cline{9-10}\cline{12-13}
&&&params&(hours)&(FLOPs)&sWUGGY (all)\textuparrow&sBLIMP\textuparrow&sSC\textuparrow&tSC\textuparrow&&perplexity\textdownarrow&auto-BLEU\textdownarrow\\\hline
\multicolumn{2}{l}{\textit{Speech-only phonetic models}}&&&&&&&&&&&\\
GSLM~\cite{lakhotia-etal-2021}${}^\dagger$      &HuBERT&&150M&6k    &$\emptyset$&64.8&54.2&53.3&66.6&&213.9&3.34\\
TWIST~\cite{NEURIPS2023_c859b99b}   &mHuBERT~\cite{NEURIPS2023_c859b99b}&&7B&155k  &2.1e21&\textbf{73.9}&59.0&55.3&74.1&&66.6&4.18\\
\hline
\multicolumn{2}{l}{\textit{Speech-only syllabic models}}&&&&&&&&&&&\\
Sylber-uLM~\cite{cho2024sylber}&Sylber&&125M&55k&$\emptyset$&$\emptyset$&60.8&$\emptyset$&$\emptyset$&&$\emptyset$&$\emptyset$\\
SyllableLM Base~\cite{baade2024}&SylBoost 6.25Hz&&85M &55k&8.2e18&72.1&62.9&$\emptyset$&70.2&&$\emptyset$&$\emptyset$\\
\hline
\multicolumn{2}{l}{\textit{Speech-text models}}&&&&&&&&&&&\\
Moshi~\cite{defossez2024moshispeechtextfoundationmodel}&Mimi~\cite{defossez2024moshispeechtextfoundationmodel}&\checkmark&7B&7M&2.9e22&72.6&58.8&60.8&83.0&&$\emptyset$&$\emptyset$\\
GLM-4-Voice~\cite{zeng2025scaling}&GLM-4-Voice-Tokenizer&\checkmark&9B&700k+&5.4e22&$\emptyset$&$\emptyset$&62.4&82.9&&$\emptyset$&$\emptyset$\\
SpiRit-LM~\cite{10.1162/tacl_a_00728}&mHuBERT&\checkmark&7B&570k&4.0e21&69.0&58.3&61.0&82.9&&62.0&4.49\\\hline
\rowcolor{cyan!10}SylReg-LM 85M (Speech-only)&SylReg-Distill&&85M&55k&8.2e18&68.3&\textbf{64.3}&54.0&71.1&&59.6&5.53\\
\rowcolor{cyan!10}SylReg-LM 85M (Speech-only)&SylReg-Distill ($C=1$)&&85M&55k&8.2e18&69.2&63.9&53.3&70.2&&59.0&5.34\\
\rowcolor{cyan!10}SylReg-LM 85M&SylReg-Distill&\checkmark&85M&55k&8.2e18&68.2&64.2&54.9&72.0&&58.4&3.80\\
\rowcolor{cyan!10}SylReg-LM 7B&SylReg-Distill&\checkmark&7B&129k&1.3e21&68.5&63.2&\textbf{67.1}&\textbf{85.4}&&\textbf{54.3}&\textbf{2.63}\\
\hline
\end{tabular}
}
\end{table*}

\subsection{Speech language modeling}
\noindent\textbf{Metrics}
We evaluate the lexical, syntactic, and semantic understanding capabilities of speech LMs using sWUGGY, sBLIMP, and StoryCloze, respectively~\cite{nguyen2020zeroresourcespeechbenchmark,NEURIPS2023_c859b99b}. All three are contrastive metrics that assess whether the model assigns higher likelihood to linguistically correct speech samples over minimally incorrect counterparts. For example, in sWUGGY, the model should score real words (e.g., ``b\underline{r}ick'') over pseudo-words (e.g., ``b\underline{l}ick''). Spoken StoryCloze (sSC) requires the model to select the correct ending for a given story and evaluates commonsense reasoning, whereas Topic StoryCloze (tSC) randomly samples incorrect endings from the dataset to assess topical coherence. Log-likelihoods are normalized by sequence length for fair scoring. To assess the generative capability, we prompt speech LMs to generate a 10-second continuation given a 3-second prefix from the LibriSpeech test-clean set. For generation, we use a softmax temperature of 0.8. We transcribe the generated speech using Whisper-large-v3~\cite{pmlr-v202-radford23a} and compute the perplexity of the transcript $\bm{w}$ with OLMo 2 1B~\cite{walsh2025}, which is independent of all LMs compared in Table~\ref{tab:gslm}. To quantify repetition in generation, we calculate auto-BLEU using 2-gram~\cite{lakhotia-etal-2021}:
\begin{align*}
\autobleu(\bm{w})=\frac{\sum_{\gamma\in\ngram(\bm{w})}\mathds{1}[\gamma\in(\ngram(\bm{w})\setminus\gamma)]}{\lvert \ngram(\bm{w})\rvert}.
\end{align*}
We also report the estimated compute in FLOPs as $6ND$, where $N$ and $D$ denote the number of non-embedding parameters and processed tokens, respectively~\cite{kaplan2020scalinglawsneurallanguage}.

\noindent\textbf{Results}
Table~\ref{tab:gslm} presents the results of speech LMs on linguistic understanding and generation. We first train a randomly initialized SyllableLM Base architecture on the speech-only Libri-Light under the same computational budget. We observe that speech-only SylReg-LM surpasses SyllableLM on high-level linguistic tasks, i.e., sBLIMP and tSC, but underperforms it on sWUGGY. This suggests that improvements in syllabic tokenization do not necessarily help speech LMs discriminate subtle phone-level distinctions between words and nonwords. While our $C=1$ variant improves sWUGGY due to its finer-grained tokens (See Bitrate in Table~\ref{tab:tts}), it deteriorates linguistically high-level syntactic and semantic metrics. As an ablation, we further initialize SylReg-LM from OPT 125M~\cite{zhang2022optopenpretrainedtransformer} and train it using speech-text interleaving. This interleaving boosts the StoryCloze scores, reflecting the benefits of knowledge transfer on semantic tasks. We also observe that the continuation quality of SylReg-LM 85M outperforms that of SpiRit-LM 7B, whereas using only 0.2\% of its compute. This high computational efficiency is partially attributed to the low frame rate of syllabic tokens. When scaling with respect to both model and data sizes, SylReg-LM 7B achieves an average relative improvement of 7\% over SpiRit-LM on sBLIMP and StoryCloze. Despite a 1-point drop on sBLIMP compared to SylReg-LM 85M, possibly due to synthetic speech in the training data~\cite{maimon-etal-2025-slamming}, the overall results demonstrate its efficacy in capturing high-level linguistic abstractions.

\begin{table}[t]
\centering
\caption{Token-to-speech resynthesis on the LibriSpeech test-clean split.}
\label{tab:tts}
\begin{tabular}{lcccc}\hline
Model       &Bitrate&CER\textdownarrow&WER\textdownarrow&UTMOS\textuparrow\\\hline
Ground truth&$\infty$&0.72&1.97&4.10\\
TWIST~\cite{NEURIPS2023_c859b99b}&174.8&2.59&5.65&3.85\\
Sylber~\cite{cho2024sylber}&$\infty$&\textbf{1.43}&\textbf{3.50}&4.13\\
SylBoost 6.25Hz~\cite{baade2024}&\textbf{75.2}&3.16&6.44&3.89\\
\rowcolor{cyan!10}SylReg-Distill (ours)&76.8&2.53&5.46&\textbf{4.31}\\
\rowcolor{cyan!10}SylReg-Distill ($C=1$)&82.3&2.91&6.06&\textbf{4.31}\\
\hline
\end{tabular}
\end{table}

\subsection{Token-to-speech resynthesis}
\noindent\textbf{Metrics}
We evaluate how faithfully speech resynthesized from syllabic tokens preserves the original spoken content. Content accuracy is measured by word error rate (WER) and character error rate (CER) between the reference text and the Whisper-large-v3 transcript, whereas perceptual quality is estimated using UTMOS~\cite{saeki22c_interspeech}. Following~\cite{NEURIPS2023_2d8911db,baade2024}, we use 4--10 second utterances from LibriSpeech test-clean for evaluation. Finally, we evaluate coding efficiency using the bitrate, defined as $(\mathrm{token\ frame\ rate})\cdot\log_2(\mathrm{vocab\ size})$.

\noindent\textbf{Results}
Table~\ref{tab:tts} summarizes the results of token-to-speech resynthesis. In Sylber, we extract speaker embeddings from a random utterance within the same speaker as the reference speech. The Sylber synthesizer achieves the lowest CER and WER, as it operates on continuous features with an unbounded bitrate. In contrast, our model achieves the highest UTMOS score. It can resynthesize clean speech from utterances with background noise such as rain, indicating its robustness to noise. Furthermore, our proposed method matches the TWIST synthesizer in CER and WER at a 2.3$\times$ lower bitrate, demonstrating its high coding efficiency. Despite its higher bitrate, a chunk size of 1 impairs intelligibility due to inaccurate syllable segmentation in Table~\ref{tab:chunk}.

\begin{table*}[t]
\centering
\caption{Impact of the merge threshold in SylReg on downstream performance.}
\label{tab:threshold}
\resizebox{\linewidth}{!}{
\begin{tabular}{lccccccccccccccccccc}\hline
\multirow{2}{*}{$\tau_\mathrm{SylReg}$}&Token&\multicolumn{7}{c}{Syllabic tokenization on the LibriSpeech test set}&&\multicolumn{6}{c}{Speech language modeling}&&\multicolumn{3}{c}{Token-to-speech resynthesis}\\\cline{3-9}\cline{11-16}\cline{18-20}
&frame rate&Pr&Re&F1&R&SP&CP&SNMI&&sWUGGY (all)&sBLIMP&sSC&tSC&perplexity&auto-BLEU&&CER&WER&UTMOS\\\hline
0.5&4.99&\textbf{71.6}&84.4&\textbf{77.5}&\textbf{76.3}&\textbf{79.6}&32.8&\textbf{91.7}&&65.8&63.1&53.6&70.3&\textbf{55.2}&9.22&&8.71&14.5&3.99\\
0.7&5.82&64.5&\textbf{88.7}&74.7&63.1&79.5&\textbf{34.9}&91.5&&\textbf{68.3}&\textbf{64.3}&\textbf{54.0}&\textbf{71.1}&59.6&\textbf{5.53}&&\textbf{2.53}&\textbf{5.46}&\textbf{4.31}\\
\hline
\end{tabular}
}
\end{table*}

\subsection{Impact of the merge threshold on downstream tasks}
Table~\ref{tab:threshold} shows the impact of the merge threshold $\tau_\mathrm{SylReg}$ on downstream tasks. This threshold controls the token frame rate. The default $\tau_\mathrm{SylReg}=0.7$ yields an oversegmented frame rate of 5.82~Hz compared to the ground-truth syllable rate of 4.38~Hz. Lowering $\tau_\mathrm{SylReg}$ to 0.5 reduces the frame rate to 4.99~Hz by further merging adjacent segments. Although coarser tokens better align with ground-truth boundaries and thus improve the R-value, which penalizes oversegmentation, they degrade downstream performance. This is particularly pronounced in sWUGGY, CER, and WER, which require fine-grained lexical discrimination. The low perplexity likely stems from reward hacking caused by word repetition, as evidenced by the high auto-BLEU score.

\section{Conclusion}\label{sec:conclusion}
We propose a self-supervised syllabic tokenization method that prioritizes linguistic content over speaker characteristics. The resulting coarse syllabic tokens alleviate the granularity mismatch between speech and text without modifying the downstream LM architecture. Experimental results show that SylReg achieves state-of-the-art performance in syllable segmentation accuracy and syllabic token quality. Moreover, SylReg-LM outperforms the phone-level token-based SpiRit-LM in syntactic and semantic understanding, highlighting the advantage of speaker-disentangled syllabic tokens for modeling high-level linguistic abstractions in speech.

Our ultimate goal is to develop intelligible and expressive spoken dialogue agents built on both phonetic and acoustic representations. This work focuses on the former, namely enhancing intelligibility, by improving the linguistic purity of syllabic tokens. Future work will explore integrating acoustic features for expressive speech generation. Moreover, SylReg can operate on continuous syllabic representations by removing the quantization module, potentially improving speech synthesis quality by avoiding the information bottleneck inherent to discrete tokenization. Finally, refining multi-stage distillation remains an important challenge for further large-scale training.

\section*{ACKNOWLEDGMENT}
We used ChatGPT and Gemini to assist with grammar correction and sentence refinement throughout the manuscript.

\bibliographystyle{IEEEtran}
\bibliography{refs}

\vfill\pagebreak

\end{document}